\documentclass[letterpaper, 10 pt, conference]{ieeeconf}  

\IEEEoverridecommandlockouts                              
\renewcommand{\thetable}{\Roman{table}}
\usepackage[utf8]{inputenc}
\usepackage[T1]{fontenc}
\usepackage{graphicx}
\usepackage{float} 
\usepackage{subfigure}
\usepackage{ragged2e}
\usepackage{booktabs}
\usepackage{bbding}
\usepackage{multirow}
\usepackage{makecell}
\usepackage[ruled,vlined]{algorithm2e}
\usepackage{amsmath,amssymb,amsfonts}
\usepackage[]{caption2}
\usepackage{cite}
\usepackage{mathptmx} 
\usepackage[colorlinks,linkcolor=blue]{hyperref}

\usepackage{fancyhdr}
\usepackage{dsfont}
\usepackage{array,color}
\usepackage{bm}
\usepackage{algpseudocode}
\usepackage{bbm}

\title{
\vspace{-0.1in}
\LARGE \bf
Opportunistic Collaborative Planning with Large Vision Model\\Guided Control and Joint Query-Service Optimization
\vspace{-0.15in}
}
\author{$^\dagger$: corresponding authors}
\author{Jiayi Chen$^{1}$, Shuai Wang$^{2,\dagger}$, Guoliang Li$^{3}$, Wei Xu$^{4}$, Guangxu Zhu$^{1,\dagger}$,
\\Derrick Wing Kwan Ng$^{5}$,~\emph{Fellow, IEEE}, and Chengzhong Xu$^{3}$,~\emph{Fellow, IEEE}
\vspace{-0.15in}
\thanks{
$^{1}$Jiayi Chen and Guangxu Zhu are with the Shenzhen Research Institute of Big Data, The Chinese University of Hong Kong (Shenzhen), Shenzhen 518115, China ({\tt\footnotesize \ jiayichen5@link.cuhk.edu.cn, gxzhu@sribd.cn}). $^{2} $Shuai Wang is with the Shenzhen Institutes of Advanced Technology (SIAT), Chinese Academy of Sciences, Shenzhen, China ({\tt\footnotesize s.wang@siat.ac.cn}). $^{3}$Guoliang Li and Chengzhong Xu are with the State Key Laboratory of Internet of Things for Smart City (SKL-IOTSC), University of Macau, Macau, China. $^{4}$Wei Xu is with the Manifold Tech Limited, Hong Kong, China. $^{5}$Derrick Wing Kwan Ng is with the School of Electrical Engineering and Telecommunications, the University of New South Wales, Australia. $\dag$ denotes corresponding authors.}
}
\begin{document}

\maketitle
\thispagestyle{empty}
\pagestyle{empty}

\begin{abstract}
Navigating autonomous vehicles in open scenarios is a challenge due to the difficulties in handling unseen objects. Existing solutions either rely on small models that struggle with generalization or large models that are resource-intensive. 
While collaboration between the two offers a promising solution, the key challenge is deciding when and how to engage the large model. To address this issue, this paper proposes opportunistic collaborative planning (OCP), which seamlessly integrates efficient local models with powerful cloud models through two key innovations.
First, we propose large vision model guided model predictive control (LVM-MPC), which leverages the cloud for LVM perception and decision making. The cloud output serves as a global guidance for a local MPC, thereby forming a closed-loop perception-to-control system.
Second, to determine the best timing for large model query and service, we propose collaboration timing optimization (CTO), including object detection confidence thresholding (ODCT) and cloud forward simulation (CFS), to decide when to seek cloud assistance and when to offer cloud service. Extensive experiments show that the proposed OCP outperforms existing methods in terms of both navigation time and success rate.
\end{abstract}
\vspace{-0.15cm}

\section{Introduction}
While autonomous navigation technologies have seen rapid advancements, achieving high efficiency under safety guarantee remains a challenge, particularly when it comes to recognizing and reacting to unknown objects in open scenarios \cite{dense-rl}. 
The presence of these objects often leads to perception errors propagating to the subsequent control, making ego-vehicle prone to take detour or get stuck \cite{han2024neupan}.

Traditional navigation systems \cite{han2023rda,han2023efficient,li2023safe} rely on small local models that struggle to handle unseen objects. Emerging large models \cite{kou2025enhancing,hu2024agentscodriver,xu2023drivegpt4} are capable of recognizing a wide variety of objects, but are resource-intensive to operate at a high frequency. While collaboration between the small and large models is a promising direction, determining when and how to engage the cloud large model is non-trivial.
First, collaboration involves interactions between large and small models. 
Feasible vehicle dynamics and collision avoidance must be ensured when black-box large models are adopted.
Second, the vehicle should decide when to seek cloud assistance based on a confidence threshold. Improper threshold may result in excessive queries or conservative local navigation. 
Third, the cloud server should decide whether to offer the inference service by comprehensively evaluating the current and future situations. 
Cloud inference may not always improve navigation performance, since the detection failures in the queried image may not be at the critical path.
Existing collaboration schemes \cite{li2024edge,zhang2024multi,wang2023bevgpt,RILaaS,sha2023languagempc,morrell2022nebula,FogROS2} fail in addressing the above challenges, since they ignore the inter-dependency between low-level motion planning and high-level large model inference and collaboration timing optimization (involving both query and service timing).

\begin{figure*}[!t]
    \centering
    \begin{minipage}[b]{0.98\textwidth} 
        \centering
        \includegraphics[width=\textwidth]{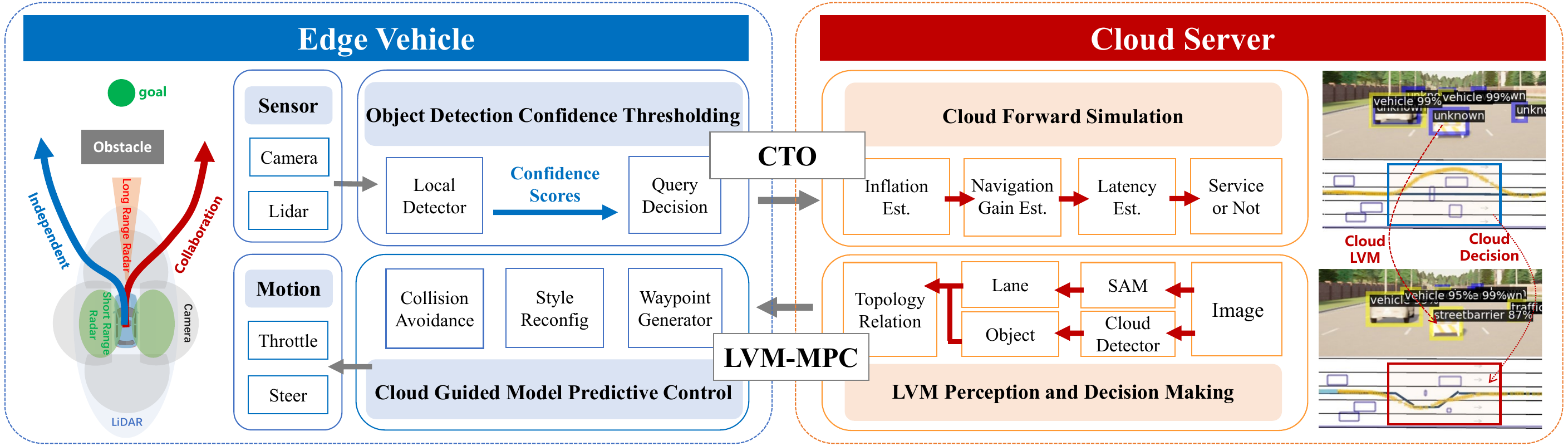}
    \end{minipage}
    \vspace{-0.05in}
    \caption{System architecture of the proposed OCP, which consists of LVM-MPC and CTO blocks.}
    \label{fig:system_overview}
    \vspace{-0.25in}
\end{figure*}

To fill the gap, this paper proposes opportunistic collaborative planning (OCP), which seamlessly integrates the efficient planning of local small models and the powerful planning of cloud large models, via two key innovations. 
The first innovation is to exploit the suitable multi-models for collaboration between cloud and edge. In particular, the cloud first executes large vision model (LVM) for perception and decision making. Its output is forwarded to the vehicle as guidance of decisions (e.g., lane change), paths (i.e., waypoints), and control styles (e.g., inflation distances), for reconfiguring the model predictive controller (MPC). This forms a closed-loop LVM-MPC collaboration from perception to control, which effectively handles rare objects in open environments.
The second innovation is the collaboration timing optimization (CTO), including both local assistance query and cloud inference service. 
Object detection confidence thresholding (ODCT) is proposed to decide when to seek cloud assistance at the local vehicle. 
Furthermore, cloud forward simulation (CFS) is proposed to decide when to offer cloud service by considering the trajectory improvements and processing latency. 

To evaluate OCP, it is necessary to adopt a high-fidelity simulator with intertwined execution of learning-based perception and optimization-based control. 
We thus implement OCP in the Car Learning to Act (Carla) platform \cite{dosovitskiy2017carla} using robot operation system (ROS).
Results in Carla-ROS demonstrate the superiority of the proposed OCP over existing RDA planner \cite{han2023rda}, and collaborative planner \cite{li2024edge}, in a multi-lane urban driving scenario. 
The success rate and navigation time are reduced by over $6\%$ and $26\%$, respectively.

Our contributions are summarized as follows:
\begin{itemize}
    \item Propose OCP to handle rare objects in open scenarios by integrating small local and large cloud models. 
    \item Propose LVM-MPC, which is a closed-loop collaboration from perception to control.
    \item Propose CTO to determine the optimal timing for large model query and inference, respectively.   
    \item Implement OCP and associated methods in Carla, and demonstrate its effectiveness in various scenarios.
\end{itemize}

\section{Related Work}
Autonomous navigation involves perception, which provides necessary information about the environment, and planning, which computes a sequence of feasible actions to reach the goal \cite{han2024neupan}. 
Conventional navigation is based on small models \cite{han2023rda,han2023efficient}. 
While easy-to-deploy, they suffer from error propagation \cite{li2023safe,zhang2024multi}: the detection results (e.g., voxels, bounding boxes) involves high uncertainties when handling unseen objects in open scenarios, leading to a larger safety distance in the planner \cite{han2024neupan}.
Large models can be adopted to mitigate error propagation. 
For instance, in \cite{kou2025enhancing}, large vision model is adopted as a backbone coupled with downstream perception head to understand street scene semantic information. 
In \cite{hu2024agentscodriver}, AgentsCoDriver is proposed, which uses large language models to enhance generalization and realize lifelong learning through cognitive memory. 
In \cite{xu2023drivegpt4}, DriveGPT4 is proposed, which employs multi-modal large models to process videos and captions, thereby effectively explaining vehicle actions. 
However, large models suffer from a lack of interpretability and the output trajectory may violate vehicle dynamics.
Due to high computation requirements, direct deployment on vehicles is not economic and may lead to low execution frequency.

Recently, the collaborative planning paradigm emerges, which deploys small efficient models at the vehicle and large powerful models at the cloud \cite{wang2023bevgpt,li2024edge}.
For instance, object recognition and grasp planning can be offloaded to cloud as a service \cite{RILaaS}.
In \cite{sha2023languagempc}, the Language-MPC is proposed to understand natural language instructions and enhance decision-making through large language models while ensuring feasible vehicle dynamics through small MPC models. 
In DARPA SubT challenge \cite{morrell2022nebula}, a collaborative planning framework is adopted to deploy the mission planner at the cloud for task assignment and the motion planner at each robot for task execution.
To reduce the communication latency introduced by collaboration, a cloud robotics platform FogROS2 is proposed to effectively connect robot systems across different physical locations, networks, and data distribution services \cite{FogROS2}. 
Nonetheless, current collaborative planning schemes ignore the collaboration timing optimization, which may lead to mismatch between large model service and small model requirement. In contrast, OCP optimizes the collaboration timing explicitly through ODCT and CFS.

\section{Opportunistic Collaborative Planning}

\subsection{Overview of Architecture}

The architecture of OCP is shown in Fig.~\ref{fig:system_overview}.
The input of OCP consists of the multi-modal sensor data (e.g., camera images, lidar point clouds) and the navigation goal position (i.e., marked as a green ball).
The output consists of collaboration states (i.e., a sequence of binary decisions), collision-free trajectories (i.e., a sequence of vehicle states), and control actions (i.e., throttle and steer). 
The goal of OCP is to transform unknown objects into known ones by calling large models at the right timing, so as to improve the vehicle trajectory quality to the maximum extent, avoiding detouring or stuck cases as shown at the right hand side of Fig.~\ref{fig:system_overview}.

To realize the above objective, the OCP system adopts an LMV-MPC block to decide how to collaborate, and a CTO block to decide when to collaborate.  
The LMV-MPC block executes LVM perception and decision making at the cloud for guiding small MPC model at the local vehicle. 
The CTO block consists of the ODCT and CFS modules, which decide when to seek cloud assistance and when to offer cloud service, respectively. 

\subsection{LVM-MPC and CTO}

For LVM-MPC, it is based on the joint processing of cloud and vehicle models. 
At the cloud, the LVM adopts segment anything model (SAM) \cite{Kirillov_2023_ICCV} for reperception, which recognizes obstacles, generates lanes, and associates obstacles with their corresponding lanes. 
With these information, the cloud further employs a decision tree for generating behavior decisions (i.e., left, right, keep) and reconfiguring control styles (i.e., inflation distances).
These information, together with the calibrated objects and new confidence scores, is forwarded to the ego vehicle. The MPC module generates responsive control commands to ensure safe efficient navigation, by considering both real-time local conditions and strategic guidance from the cloud. 
Under the non-collaboration mode, the MPC is able to operate independently based on local information.

For CTO, the ODCT module processes vehicle images, assigning confidence scores to detected objects. Based on confidences, it outputs a cloud query decision to determine whether to proceed independently or seek cloud assistance. If ODCT finds low confidence objects, it would upload the associated detection results and confidence scores to the cloud. 
Upon receiving the vehicle query, the CFS assesses the necessity for collaboration by jointly considering potential trajectory improvements and processing delays.
If CFS accepts the query, LVM-MPC will be initiated, and the vehicle would subsequently upload streaming images and point clouds to the cloud server. 
Otherwise, the vehicle keeps on local navigation.

\section{LVM Guided MPC}

The proposed LVM-MPC is designed to find a collision-free trajectory under the guidance of LVM and constraints of vehicle dynamics. 
The system operation is divided into $T$ time slots, and the duration between consecutive slots is $\Delta t$.
At the $t$-th time slot, $t \in\{1, \ldots, T\}$, the input image is concatenated into a vector $\mathbf{x}_t$. 
The vehicle state is ${\mathbf{s}}_t=({a}_{t},{b}_{t},{\theta}_{t})$, where $(a_t,b_t)$ and $\theta_t$ denote position and orientation, respectively.
The control command is $\mathbf{u}_t=(v_t,\phi_t)$, with $v_t$ and $\phi_t$ representing linear velocity and steering angle, respectively. 
The collaboration state is $\beta_t\in\{0,1\}$.

The consecutive states, $\mathbf{s}_{t}$ and $\mathbf{s}_{t + 1}$, must adhere to the discrete-time kinematic model:
\begin{align}\label{eq1}
{\mathbf{s}_{t + 1}} = {\mathbf{s}_t} + f({\mathbf{s}_t},{\mathbf{u}_h})\Delta t,
\end{align}
where $f(\cdot)$ is the state evolution function determined by the vehicle dynamics \cite{han2023rda}. 
The motion operation has its physical limits, which is given by
\begin{align}
\mathbf{u}_{\text{min}} \le \mathbf{u}_{t} \le \mathbf{u}_{\text{max}}, 
\end{align}
where $\mathbf{u}_{\text{min}}$ and $\mathbf{u}_{\text{max}}$ are the minimum and maximum input limits, respectively. 
To prevent harsh braking or acceleration and improve the vehicle energy consumption, we must also control the change rate of motions:
\begin{align}\label{eq3}
\Delta \mathbf{u}_{\text{min}} \le \mathbf{u}_{t} - \mathbf{u}_{t-1} \le \Delta \mathbf{u}_{\text{max}},
\end{align}
where $\Delta \mathbf{u}_{\text{min}}$ and $\Delta \mathbf{u}_{\text{max}}$ are the minimum and maximum bounds on change rates of motion vectors.

\subsection{LVM Perception}

If $\beta_t=0$, the vehicle adopts a local detection model, denoted as a function $g_{\mathrm{local}}(\cdot)$, to map $\mathbf{x}_t$ into a list ${\mathcal{B}}_{t}=g_{\mathrm{local}}(\mathbf{x}_{t})$, where $\mathcal{B}_{t}= \{(y_{1,t},\mathbb{O}_{1,t}),\cdots,(y_{M,t},\mathbb{O}_{M,t})\}$, with $y_{m,t}$ and $\mathbb{O}_{m,t}$ representing the category and bounding box of the $k$-th obstacle. 
Each box is assigned a confidence level $c_{m,t} \in [0,1]$. 
The design of the local detection model is based on the unsniffer architecture \cite{liang2023unknown}.

With unknown objects in $\mathbf{x}_t$, $\mathcal{B}_{t}$ may mismatch from the ground truth $\mathcal{B}_{t}^*$, forcing the vehicle to take a detour. 
To this end, we need to shift the collaboration state from $\beta_t=0$ to $\beta_t=1$.
Then, the cloud server adopts an LVM-based detection model, denoted as a function $g_{\mathrm{cloud}}(\cdot)$, to calibrate $\mathcal{B}_{t}$ into $\mathcal{B}_{t}^\diamond=
g_{\mathrm{cloud}}(\mathbf{x}_t)$.

\begin{figure}[!t]
    \centering
    \includegraphics[width=0.9\columnwidth]{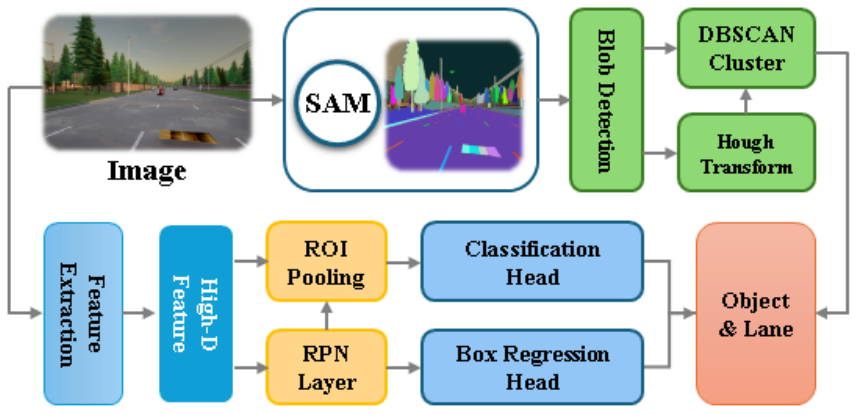}
    \vspace{-0.05in}
    \caption{LVM cloud perception based on SAM.}
    \label{figure:sam}
    \vspace{-0.2in}
\end{figure}

The architecture of our cloud detection model is illustrated in Fig.~\ref{figure:sam}. First, SAM \cite{Kirillov_2023_ICCV,ravi2024sam2} generates a comprehensive set of binary masks $\{\mathbf{v}_i\}$ through zero-shot segmentation, where each mask $\mathbf{v}_i$ corresponds to diverse image elements including objects, lanes, background regions, and edge contours. The mask representation is formally defined as:
\begin{align}
\mathbf{v}_i(p) = \begin{cases}
1, & \text{if pixel } p \text{belongs to a segmented region} \\
0, & \text{otherwise}
\end{cases}. \nonumber
\end{align}

Second, we localize the road surface by identifying $\mathbf{v}_{\text{road}}$---the mask with maximal pixel continuity along the image's bottom edge, leveraging the prior that roads typically extend toward the camera viewpoint. Third, a proximity-based filter isolates traffic-relevant masks: for each candidate $\mathbf{v}_i$, we retain it only if there exists a pixel $p \in \mathbf{v}_i$ satisfying
\begin{align}
    \min_{p' \in \mathbf{v}_{\text{road}}} \|p - p'\|_2 < \delta_{\text{pixel}},
\end{align}
where $\delta_{\text{pixel}}$ is set to 0.01\% of the image width to tolerate minor localization errors. Finally, density-based clustering merges fragmented SAM regions within unsniffer's bounding boxes $B_t^\diamond$, ensuring spatial consistency between segmentation masks and detection outputs while recalibrating confidence scores $\{c_{m,t}^\diamond\}$ through cross-modal alignment.

To generate the behavior decision $\gamma_t\in\{\mathrm{keep}, \mathrm{left},\mathrm{right} \}$, the key is to 
obtain the object-lane topology relation by associating each object with a lane ID.
Here we propose to detect lane markings by checking the aspect ratio and orientation of each SAM mask.
Specifically, we introduce the concept of slope difference between the left and right edges of the mask. 
Given the width \( W \) of the segment rectangle and the camera intrinsics \( V \) (height) and \( \omega \) (rotation angle), the slope difference between the left and right edges of the mask in the image plane is given by:
\begin{align}
    \Delta\,\text{slope} = \frac{W}{V \cos\omega}.
\end{align}
This value helps us to filter out non-lane marking objects by comparing $\Delta\,\text{slope}$ for each mask with target $\Delta\,\text{slope}$ for lane markings. 
As such, the objects in $\mathcal{B}_{t}^\diamond$ are associated with the lane IDs. 
The fused information, including object boxes, object categories, lane positions, and object-lane topology relations, are passed into a decision tree, which generates behavior decision $\gamma_t$ for subsequent planning and control. 

\subsection{Cloud Guided MPC}

The MPC module is required to operate in a receding-horizon fashion that allows repeated planning at a high frequency.
We generate a local horizon that is divided into $H$ time slots with $\mathcal{H}=\{t,\cdots,t+H\}$, where $H$ is the length of receding horizon. 
Let $\mathcal{S}=\{\mathbf{s}_t,\cdots,\mathbf{s}_{t+H}\}$ and $\mathcal{U}=\{\mathbf{u}_t,\cdots,\mathbf{u}_{t+H}\}$ denote the states and actions to be optimized within the local horizon. 
First, they need to satisfy vehicle dynamic constraints \eqref{eq1}--\eqref{eq3}, which are expressed as $\{\mathcal{S},\mathcal{U}\}\in\mathcal{F}$.
Second, they need to satisfy collision avoidance constraints.
Given the local perception $\mathcal{B}_{t},\{c_{m,t}\}$, cloud perception $\mathcal{B}_{t}^\diamond,\{c_{m,t}^\diamond\}$, and the collaboration state $\beta_t$, the collision avoidance constraints are written as \cite{zhang2024multi,han2023rda}
\begin{align}
&\min_{\mathbf{v},\mathbf{o}} \{\|\mathbf{v}-\mathbf{o}\|_2| \mathbf{v}\in\mathbb{V}_{h}(\mathbf{s}_h), \mathbf{o}\in (1-\beta_t)\mathbb{O}_{m,h} + \beta_t\mathbb{O}_{m,h}^\diamond \}
\nonumber\\
& \geq 
d_0 + (1-\beta_t)(1-c_{m,h})d_{\text{inf}}
+\beta_t(1-c_{m,h}^\diamond)d_{\text{inf}}
,~\forall m,h, \label{collision2}
\vspace{-0.15in}
\end{align}
where $\mathbb{V}(\mathbf{s}_{h}) $ denote the real-time vehicle moving polytope,
$d_0$ is the minimum safety distance, $d_{\text{inf}}$ is the maximum inflation distance. 
As such, a larger safety distance is guaranteed when the confidence score is smaller \cite{zhang2024multi, li2023safe}. 
Moreover, collaboration state $\beta_t$ would also impact the inflation distance, as it switches between the confidence scores $c_{m,h}$ and $c_{m,h}^\diamond$.
Lastly, reconfiguring the inflation distance would also \emph{reconfigure the control style, e.g., from conservative to aggressive control}.

Having the vehicle dynamics and collision avoidance constraints satisfied, it is then crucial to design a cost function to optimize the navigation performance. 
In particular, given the vehicle decision $\{\gamma_t\}$, the cloud decision $\{\gamma_t^\diamond\}$, and the collaboration state $\beta_t$, the guidance waypoints are generated from the target lane ID as
$\mathcal{W}^\diamond=\{\mathbf{w}_{t}^\diamond,\mathbf{w}_{t+1}^\diamond,\cdots,\mathbf{w}_{t+H}^\diamond\}$ if $\beta_t=1$ and 
$\mathcal{W}=\{\mathbf{w}_{t},\mathbf{w}_{t+1},\cdots,\mathbf{w}_{t+H}\}$ if $\beta_t=0$.\footnote{In our experiment, the specific values of $\mathcal{W}^\diamond,\mathcal{W}$ are extracted from the high-definition map provided by the Carla simulator.}
As such, the cost function of MPC is given by the distance between the guidance and executable trajectories:
\vspace{-0.05in}
\begin{align}
    C(\mathcal{S}) &= \sum_{h=t}^{t+H} \left\| \mathbf{s}_{h} -\left[
    (1 - \beta_t)\mathbf{w}_{t} + \beta_t \mathbf{w}_{t}^{\diamond} \right]
    \right\|^2.
    \vspace{-0.05in}
\end{align}
It can be seen that the collaboration states would impact the reference waypoints.
Based on above discussions, the cloud guided MPC problem is formulated as
\begin{align}
\label{Plocal}
\mathsf{P}:\min_{\substack{\{\mathcal{S},\,\mathcal{U}\}\in\mathcal{F}}}~C(\mathcal{S}), \quad \text {s.t.}~\eqref{collision2},
\vspace{-0.1in}
\end{align}
which is solved via the RDA method in real time \cite{han2023rda}.

\section{Collaboration Timing Optimization}

For LVM-MPC in Section IV, the key is to determine the collaboration states $\{\beta_t\}$. 
Improper collaboration may not improve navigation performance while involving additional computation (e.g., large model inference) and communication (e.g., sensor data sharing) costs.
To determine the best collaboration timing, we need joint query-service optimization, where the vehicle adopt ODCT to decide when to seek cloud assistance and the cloud adopts CFS to decide whether to offer cloud service.

\subsection{Large Model Query Optimization via ODCT}

As mentioned in Section IV-A, the local detection would output confidence values $\{c_{m,t}\}$ for all detected boxes. 
The ODCT needs to decide whether to trigger a large model query, denoted as $\alpha_t\in\{0,1\}$, to the cloud for assistance based on these $\{c_{m,t}\}$. 

Here, we adopt a simple approach termed threshold based decision, which classify objects as either known or unknown by comparing a confidence threshold, \( C_{\text{threshold}} \), with the confidence score \( c_{m,t} \) of a bounding box. If \( c_{m,t} \) is below \( C_{\text{threshold}} \), the object is considered unknown. 
If there exists any unknown object, a query decision, i.e., \( \alpha_t=1 \), is triggered, prompting the vehicle to upload current detection results to the cloud for CFS.

To set a proper \( C_{\text{threshold}} \), we propose the ODCT algorithm, which optimizes \( C_{\text{threshold}} \) by splitting the data into a training dataset $\mathcal{D}_{\text{train}} = \{ (\mathbf{x}_i, y_i) \}_{i=1}^{N_{\text{train}}}$ and validation dataset $\mathcal{D}_{\text{val}} = \{ (\mathbf{x}_j, y_j) \}_{j=1}^{N_{\text{val}}}$. 
Let $\mathcal{Y}_{\text{train}}$ denote the set of known class labels.
The training dataset contains only known objects, i.e., $\forall y_j \in \mathcal{Y}_{\text{train}}$  while the validation dataset contains both known and unknown objects, i.e., $\exists y_j \notin \mathcal{Y}_{\text{train}}$.
By training a model on $\mathcal{D}_{\text{train}}$ and validating the model on 
$\mathcal{D}_{\text{val}}$, we can obtain the predicted class $\hat{y}_j$ of the $j$-th object and its associated confidence $c_j$.
With $\{\hat{y}_j,c_j\}$, we then optimize \( C_{\text{threshold}} \) by maximizing the objective function \( G(C_{\text{threshold}}) \) over the validation set as 
\begin{align}\label{threhold opt}
    &
    C_{\text{threshold}}^* = \arg\max_{C_{\text{threshold}\in[0,1]}} G(C_{\text{threshold}}),
\end{align}
where
\begin{align}\label{recall}
& G(C_{\text{threshold}}) = 
\underbrace{\frac{\sum_{j=1}^{N_{\text{val}}} \mathbb{I}(\hat{y}_j \notin \mathcal{Y}_{\text{train}}) \cdot \mathbb{I}(c_j < C_{\text{threshold}})}{\sum_{j=1}^{N_{\text{val}}} \mathbb{I}(\hat{y}_j \notin \mathcal{Y}_{\text{train}})} }
_{recall~of~correct~unknown~detections}
\nonumber\\
&\quad\quad\quad+\underbrace{\frac{\sum_{j=1}^{N_{\text{val}}} \mathbb{I}(\hat{y}_j \in \mathcal{Y}_{\text{train}}) \cdot \mathbb{I}(c_j > C_{\text{threshold}})}{\sum_{j=1}^{N_{\text{val}}} \mathbb{I}(\hat{y}_j \in \mathcal{Y}_{\text{train}})}}
_{recall~of~correct~known~detections}
\end{align}
and $\mathbb{I}$ is indicator function.
It can be seen that $J(C_{\text{threshold}})$ automatically balances the detection of unknown objects, while minimizing false positives from known objects. 
This $G$ function is guaranteed to be unimodal, since the number of correct unknown detections is monotonically increasing in $C_{\text{threshold}}$ and the number of correct known detections is monotonically decreasing in $C_{\text{threshold}}$.

\subsection{Large Model Service Optimization via CFS}

The CFS aims to balance trajectory improvement against cloud query costs by simulating agent interactions in the environment \cite{ding2021epsilon}\cite{lauri2016planning}. 
It takes local detections $\mathcal{B}_t$, local confidence $\{c_{m,t}\}$, and vehicle kinematics in \eqref{eq1}--\eqref{eq3} as input and generates the service decision $\beta_t$ as output. 
At this pre-collaboration point, the cloud has not received the image $\mathbf{x}_t$ yet and the cloud perception results $\mathcal{B}_t^\diamond,\{c_{m,t}^\diamond\}$ are not known.
Therefore, we need to predict what will happen after the large model engagement. 

Specifically, let function $J(\{c_{m,t}\},H)$ denote the trajectory length of $H$ time slots given detection confidences $\{c_{m,t}\}$. 
To compute $J$, we need to solve problem \eqref{Plocal} and obtain the optimal solution $\mathcal{S}^*=\{\mathbf{s}_t^*(\{c_{m,t}\}),\cdots,\mathbf{s}_{t+H}^*(\{c_{m,t}\})\}$, where the optimal state $\{\mathbf{s}_t^*\}$ is a function of $\{c_{m,t}\}$ due to constraint \eqref{collision2}.
The trajectory length under $\beta_t=0$ is
\begin{align}
J(\{c_{m,t}\},H)=\sum_{i=t}^{t+H-1}\|\mathbf{s}_{i+1}^*(\{c_{m,t}\})-\mathbf{s}_i^*(\{c_{m,t}\})\|,
\end{align}
which completes the MPC forward simulation once. 

Now, assume that the confidences after cloud perception follow a certain distribution $c_{m,t}^\diamond\in \mathcal{P}$. 
For example, we can adopt uniform distribution $\mathcal{P}=\mathcal{U}(C_{\text{threshold}}-\Delta c, C_{\text{threshold}}+\Delta c)$, where the mean is set to $C_{\text{threshold}}$ since the LVM is likely to recognize the objects, transforming unknown into known ones, and $\Delta c$ (e.g., $\Delta c=0.1$) accounts for the failure detection cases of LVM. 
Therefore, the expected trajectory improvement achieved by redetecting obstacles via collaboration is given by 
\begin{align} \label{improve}
\mathbb{E}[\Delta J] &= 
\int_{c_{m,t}^\diamond\in\mathcal{P}}\left[J(\{c_{m,t}^\diamond\},H)-J(\{c_{m,t}\},H)\right] d c_{m,t}^\diamond
\nonumber\\
&\approx \frac{1}{N}
\sum_{\{c_{m,t}^{(n)}\}} \left[J(\{c_{m,t}^{(n)}\},H)-J(\{c_{m,t}\},H)\right],
\end{align}
where the second line of approximation is obtained by collecting $N$ samples from $\mathcal{P}$, resulting in $N$ times of MPC forward simulations.

In practice, there may exist communication delay $T_{\text{com}}$ and computation delay $T_{\text{comp}}$ during collaboration and the trajectory improvement in \eqref{improve} may be degraded. 
In such a case, the vehicle would first execute local navigation for a time duration of $T_{\text{com}}+T_{\text{comp}}$, which correspond to $L=\lfloor(T_{\text{com}}+T_{\text{comp}})/\Delta t \rfloor$ time slots, and then execute collaborative navigation for $H-L$ time slots.
Thus, the expected trajectory length under latency is
\begin{align}
&J'(\{c_{m,t},c_{m,t}^\diamond\},L,H)=\sum_{i=t}^{t+L-1}\|\mathbf{s}_{i+1}^*(\{c_{m,t}\})-\mathbf{s}_i^*(\{c_{m,t}\})\|
\nonumber\\
&\quad\quad+
\sum_{i=L}^{t+H-1}\|\mathbf{s}_{i+1}^*(\{c_{m,t}^\diamond\})-\mathbf{s}_i^*(\{c_{m,t}^\diamond\})\|.
\end{align}
Accordingly, the expected trajectory improvement becomes 
\begin{align}
\mathbb{E}[\Delta J'] &= 
\int_{c_{m,t}^\diamond\in\mathcal{P}}\left[J'(\{c_{m,t},c_{m,t}^\diamond\},L,H)-J(\{c_{m,t}\},H)\right] d c_{m,t}^\diamond. \nonumber
\end{align}

\begin{figure}[!t]
    \centering
    \includegraphics[width=0.8\columnwidth]{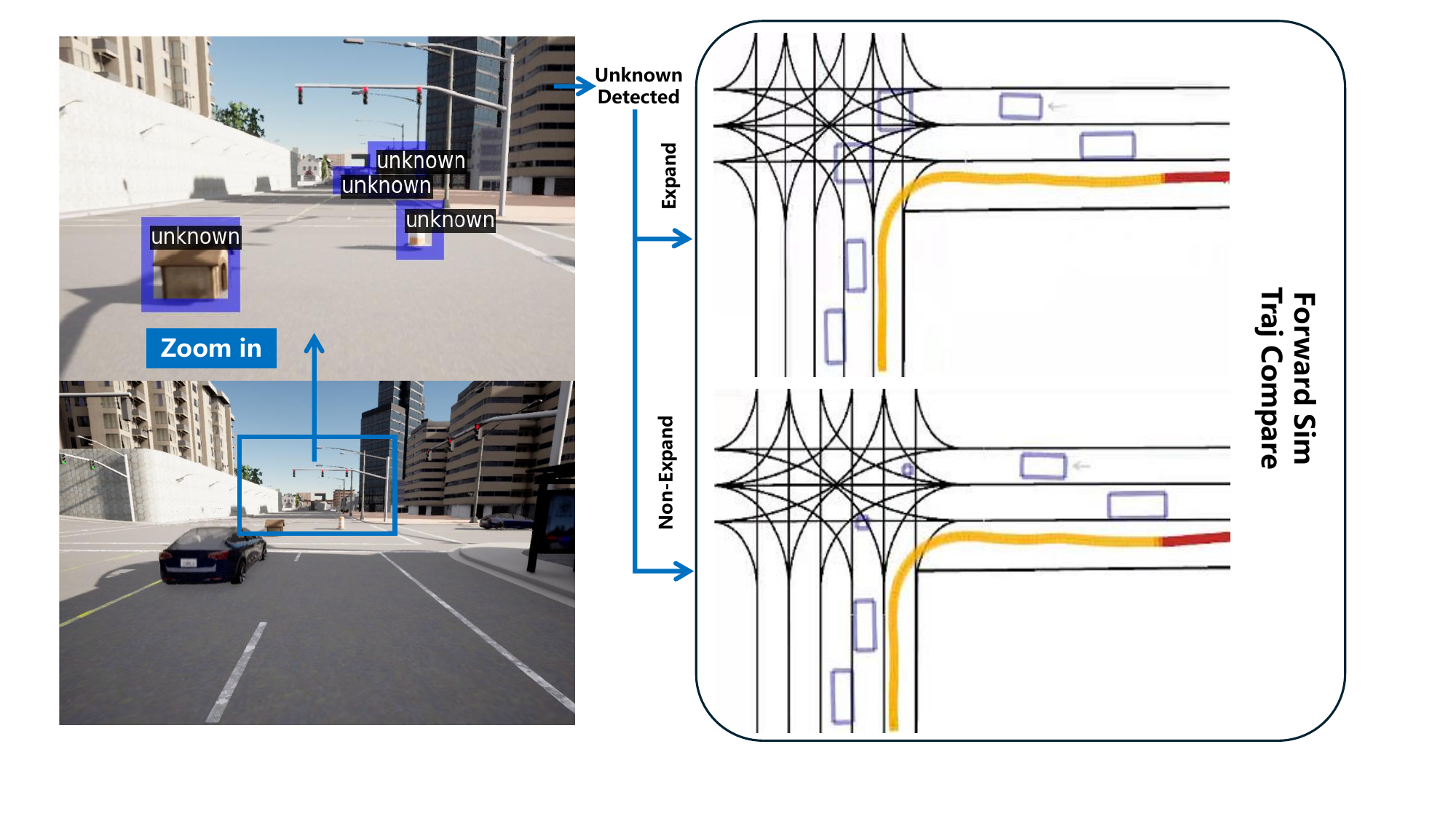}
    \vspace{-0.1in}
    \caption{The case of $\alpha_t=1$ but $\beta_t=0$ for CFS.}
    \label{figure:cfs}
    \vspace{-0.25in}
\end{figure}

To understand how CFS works, an Carla experiment is illustrated in Fig.~\ref{figure:cfs}. 
The ego vehicle detects unknown objects and trigger the cloud query, i.e., $\alpha_t=1$. 
However, the cloud finds no significant trajectory change after CFS. 
Consequently, it will rejects the query and outputs $\beta_t=0$, making the ego vehicle to navigate on its own.

\section{Experiments}

We implemented the OCP system using Python and ROS in the CARLA platform \cite{dosovitskiy2017carla}. 
The ego-vehicle, a Tesla Model3 with longitudinal and lateral wheelbases of $2.87$\,m and $1.75$\,m, is equipped with an RGBD camera and a $64$-line lidar at $10\,$Hz.
The local detection model processes sensory inputs at up to 100\,FPS and local MPC model generates control outputs at 20\,Hz. 
The safety distance of MPC is $d_0=0.3$\,m and $d_{\text{inf}}=3.0$\,m.
The length of prediction horizon is $H=18$, with time step of $\Delta t = 0.3$\,s. 
All experiments are conducted on a Ubuntu workstation with two NVIDIA RTX 3090 GPUs. 
For comparison, we implement the following schemes: 1) \textbf{Local-only strategy (LOS)}, which adopts local perception and a state-of-the-art MPC (i.e., RDA) \cite{han2023rda}; 2) \textbf{Periodical collaboration strategy (PCS)}: which is a baseline collaboration scheme with fixed interval between consecutive cloud services \cite{RILaaS};
3) \textbf{OCP (ours)}, which is the LVM-MPC collaboration scheme based on ODCT and CFS. 
Note that the collaborative planner in \cite{li2024edge} does not involve LVM, and has a similar performance as LOS \cite{han2023rda}.

\vspace{-0.05in}

\subsection{Experiment 1: Unknown Object Detection}

We first conduct experiments to validate the LVM perception and ODCT.
We collect a large dataset in Carla Town06 map, generating over 30,000 frames of 1080x720 RGB images synchronized with vehicle odometries.
Each frame is annotated with ground truth bounding boxes and undergoes rigorous quality filtering to ensure object visibility, with occluded or partially visible instances excluded during data curation.
As shown in Table. \ref{tab:data_splits}, the dataset contains 49 object categories (including vehicles and 48 other objects in Carla) and partitioned into three subsets for open vocabulary detection. 
Out of the $49$ categories, we set $38$ categories as known and the remaining $11$ as unknown.
The training set of 9,647 samples covers all 38 known classes.
The validation set contains 2,060 samples with 21 known classes and 5 unknown classes. 
The test set contains 3,400 samples with 19 known classes and the other 6 unknown classes.\footnote{The test set involves objects such as trashcan, doghouse and streetbarrier. This  aims to demonstrate the robustness of the recognition system in real-world settings. In the subsequent experiments 2--4, all the unknown objects are drawn from the test set.}
Unknown classes in validation and test sets are strictly non-overlapping and excluded from the training subset.

The local and cloud models are implemented based on the unsniffer architecture \cite{liang2023unknown}, which provides confidence score for each detection. 
The local detection model is trained on the standard training subset (9,647 samples) and fine-tuned on the validation set to optimize $C_{\text{threhold}}$. 
The final evaluation is performed on the test set. 
A complete training set of 15,107 samples is adopted that spans the 49 classes to train the cloud detection model.

\begin{figure}[t]
    \centering
    \subfigure[Validation set.]{
    \label{fig:base_scena_val}
    \includegraphics[width=0.23\textwidth]{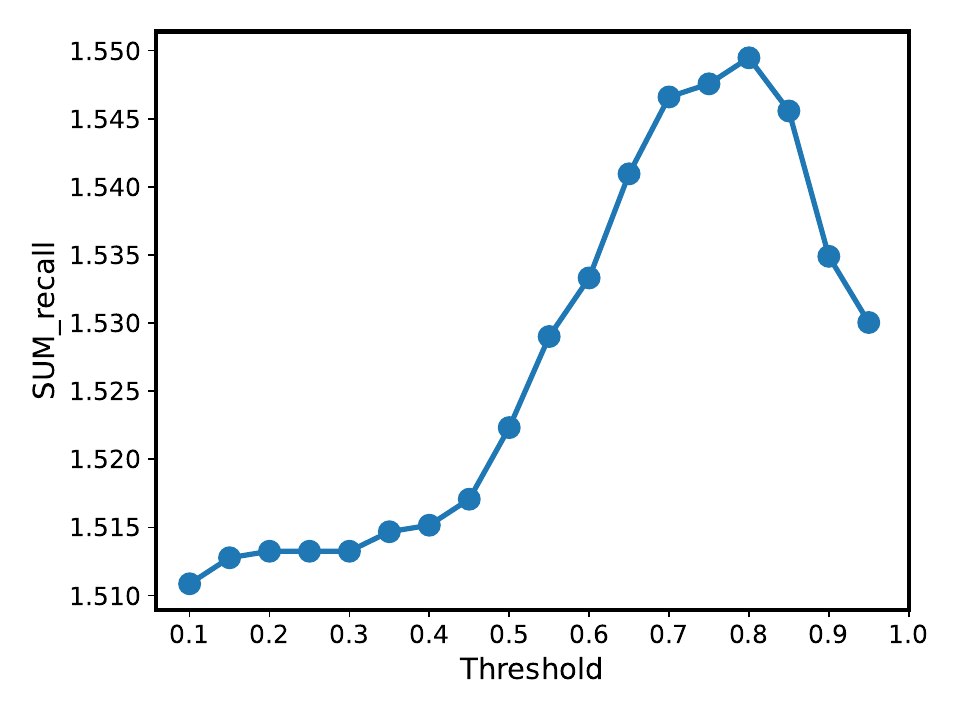}}
    \subfigure[Test set.]{
    \label{fig:base_scena_test}
    \includegraphics[width=0.23\textwidth]{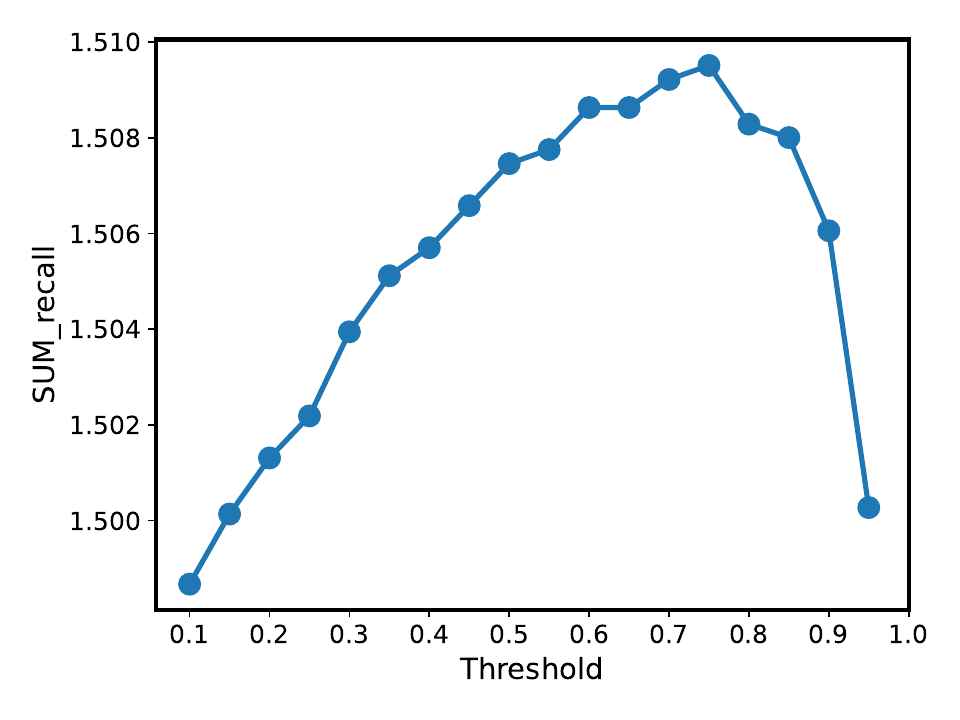}}
    \vspace{-0.15in}
    \caption{Experimental results of ODCT.}
    \label{fig:threshold_optimize}
    \vspace{-0.15in}
\end{figure}

First, the value of $G$ (i.e., sum of two recalls) in \eqref{recall} versus \( C_{\text{threshold}} \) is shown in Fig. \ref{fig:threshold_optimize}. It can be seen from Fig. \ref{fig:threshold_optimize}a that the recall score reaches its maximum on the validation set when \( C_{\text{threshold}} = 0.8 \) . 
As shown in Fig. \ref{fig:threshold_optimize}b, the highest recall score on the test set is achieved at \( C_{\text{threshold}} = 0.75 \).
This implies that the threshold optimized on the validation set exhibits strong generalization to the test set, although the two sets contain non-overlapping unknown classes.
The threshold-optimized local model achieves recalls of 0.83 (known) and 0.71 (unknown) on the validation set, with test performance showing improved known-class recall (0.89) but degraded unknown detection (0.61). The cloud model, trained on all predefined classes (no "unknown" category), attains 0.89 recall on the known-class validation set.

\begin{table}[!t]
\centering
\caption{Dataset splits (known vs. unknown classes)}
\vspace{0.05in}
\label{tab:data_splits}
\scalebox{1.1}{
\begin{tabular}{lcccc}
\hline
\textbf{Split} & \textbf{\# Samples} & \textbf{Known} & \textbf{Unknown} & \textbf{Total} \\ 
\hline
$Local_{Train}$ & 9,647 & 38 & 0 & 38 \\
$Local_{Val}$   & 2,060   & 21 & 5 & 26 \\
$Local_{Test}$  & 3,400   & 19  & 6 & 25 \\
$Cloud_{Train}$ & 15,107 & 49 & 0 & 49 \\
\hline
\end{tabular}
}
\vspace{-0.15in}
\end{table}

\begin{figure}[!t]
    \centering
    \begin{minipage}[b]{0.45\textwidth} 
        \centering
        \includegraphics[width=\textwidth]{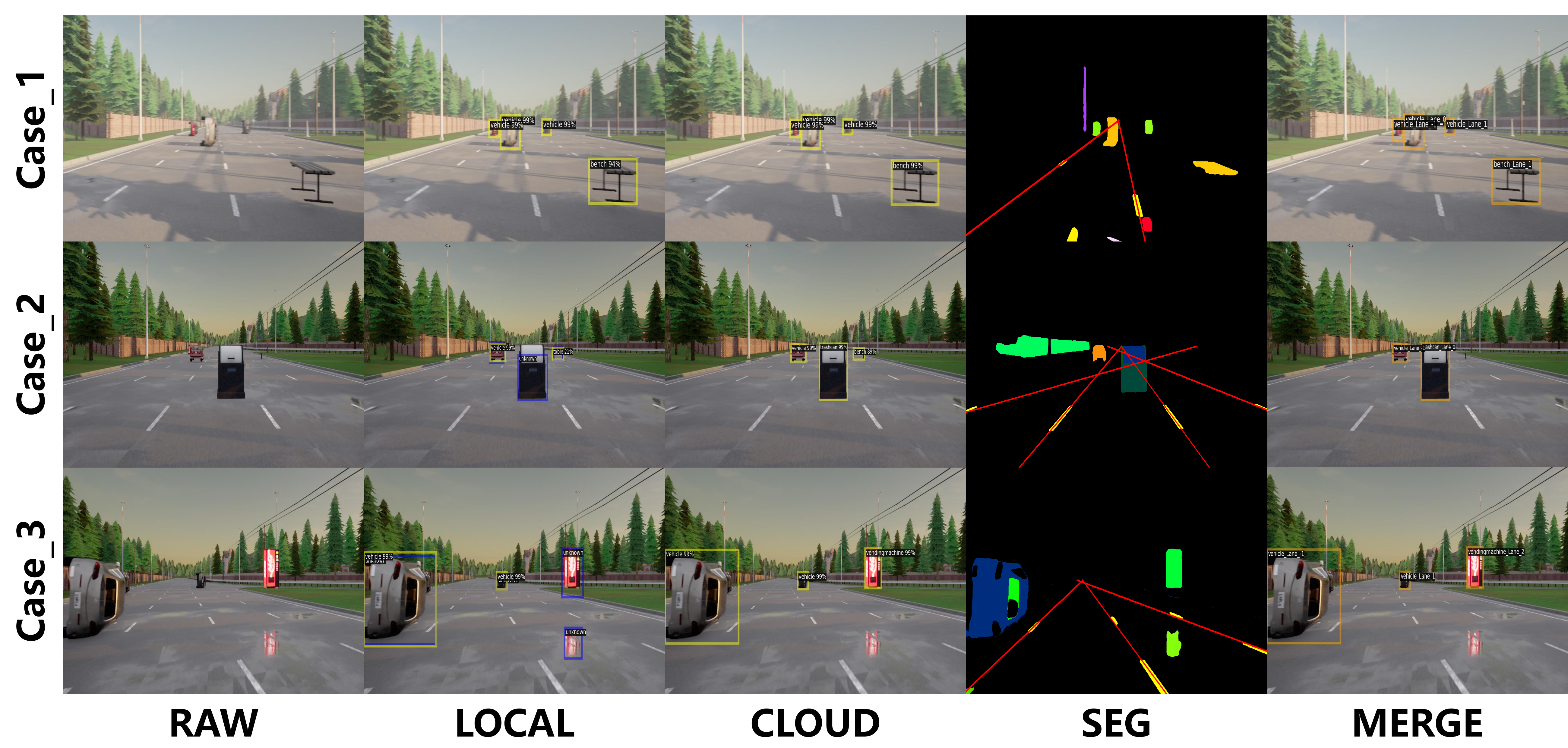}
    \end{minipage}
    \vspace{-0.1in}
    \caption{Comparison of local and cloud perception models.}
    \label{fig:show_result_2}
    \vspace{-0.2in}
\end{figure}

Next, we evaluate the perception performance of the local and cloud models. 
The associated detection results of the cloud and local models are shown in Fig.~\ref{fig:show_result_2} (the images are zoomed in for better clarity).
We consider 3 cases: case 1 contains only known objects while cases 2--3 contain unknown objects.
First, it can be seen that the local model successfully generates bounding boxes for all the known and unknown objects in cases 1--3. 
However, it does not recognize the categories of unknown objects (i.e., trash can and vending machine) in cases 2 and 3.
Fortunately, by leveraging unsniffer, the local model generates low confidence scores below the threshold \( C_{\text{threshold}}=0.8\) for these objects.
This prompts the vehicle to generate an ``unknown object'' warning, triggering a cloud query. 
Second, the cloud model, with its superior intelligence, correctly recognizes the categories of the unknown objects. 
The cloud model successfully segments the image into lane (yellow mask) and object masks, and merges the segments with detections for generating the object-lane topology relation (i.e., the lane ID of each object as shown in the MERGE column in Fig.~\ref{fig:show_result_2}).

\begin{figure}[!t]
    \centering
    \includegraphics[width=0.9\columnwidth]{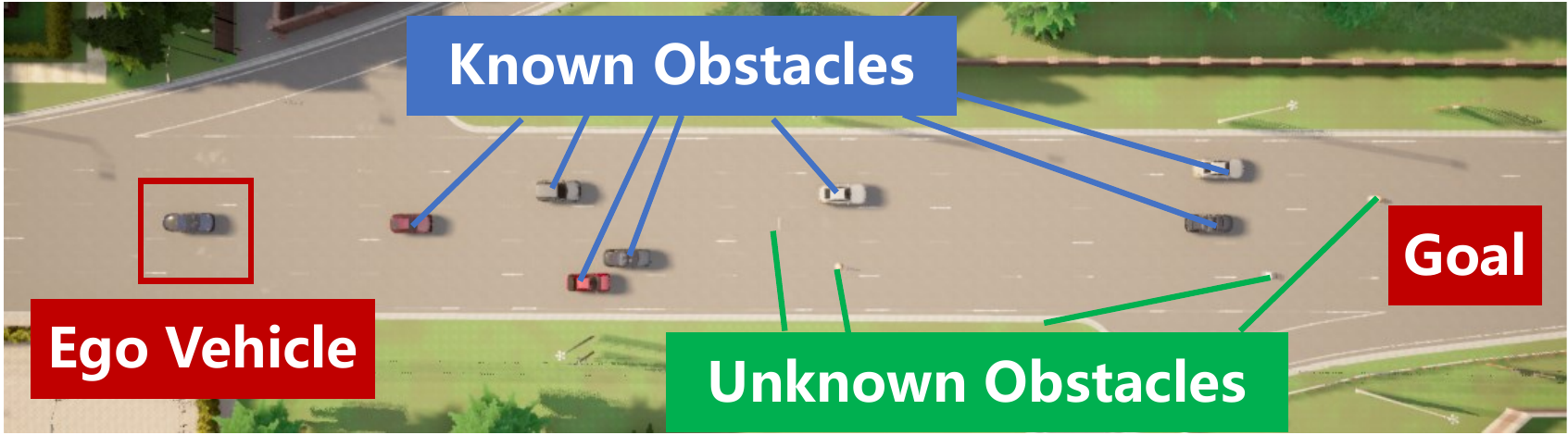}
    \vspace{-0.1in}
    \caption{Scenario configuration for experiment 2.}
    \vspace{-0.2in}
    \label{figure:setting}
\end{figure}

\begin{figure*}[t]
    \centering
    \vspace{-0.2cm}
    \subfigure[LOS]{
    \label{fig:base_scena}
    \includegraphics[width=0.3\textwidth]{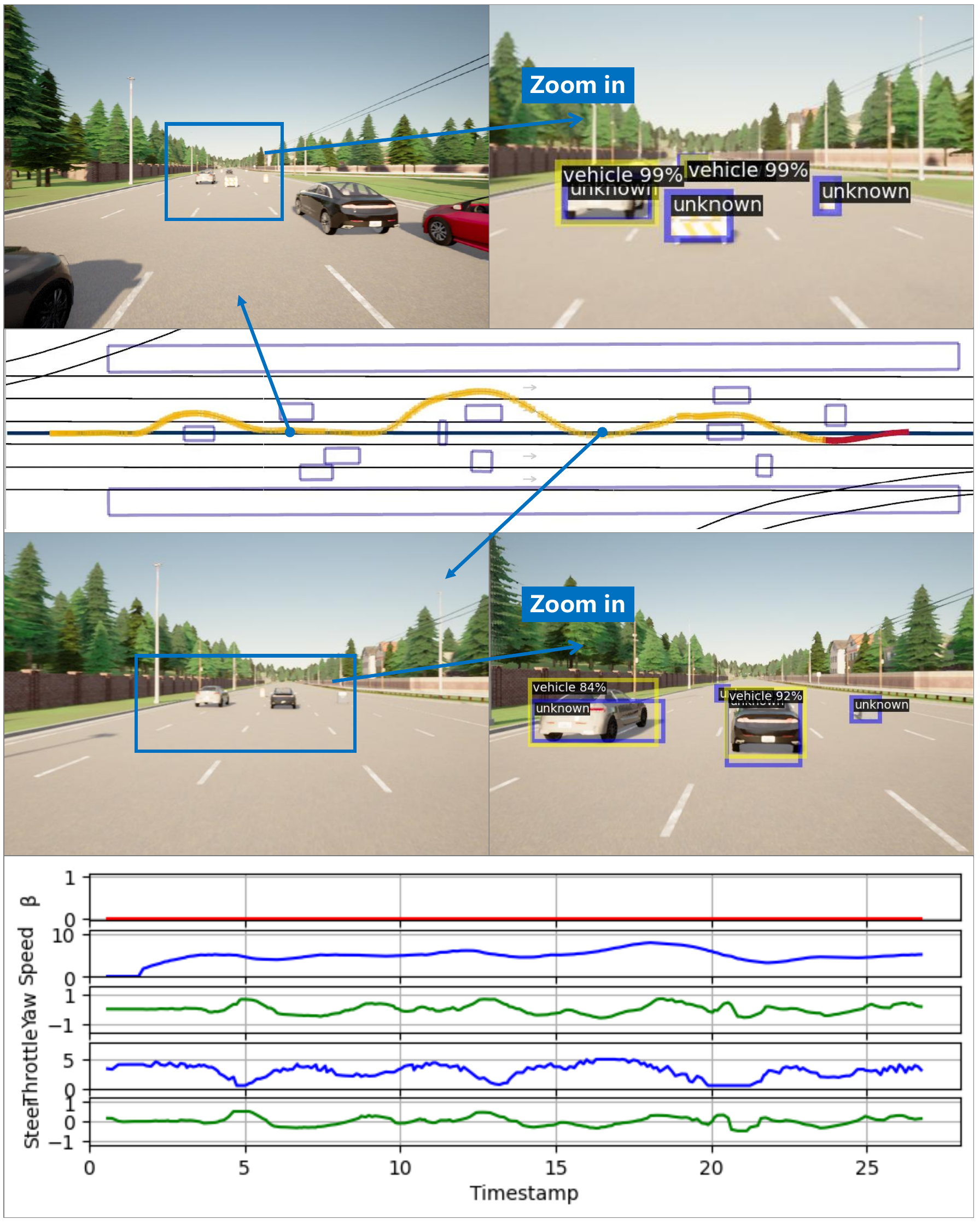}}
    \subfigure[PCS]{
    \label{fig:local}
    \includegraphics[width=0.3\textwidth]{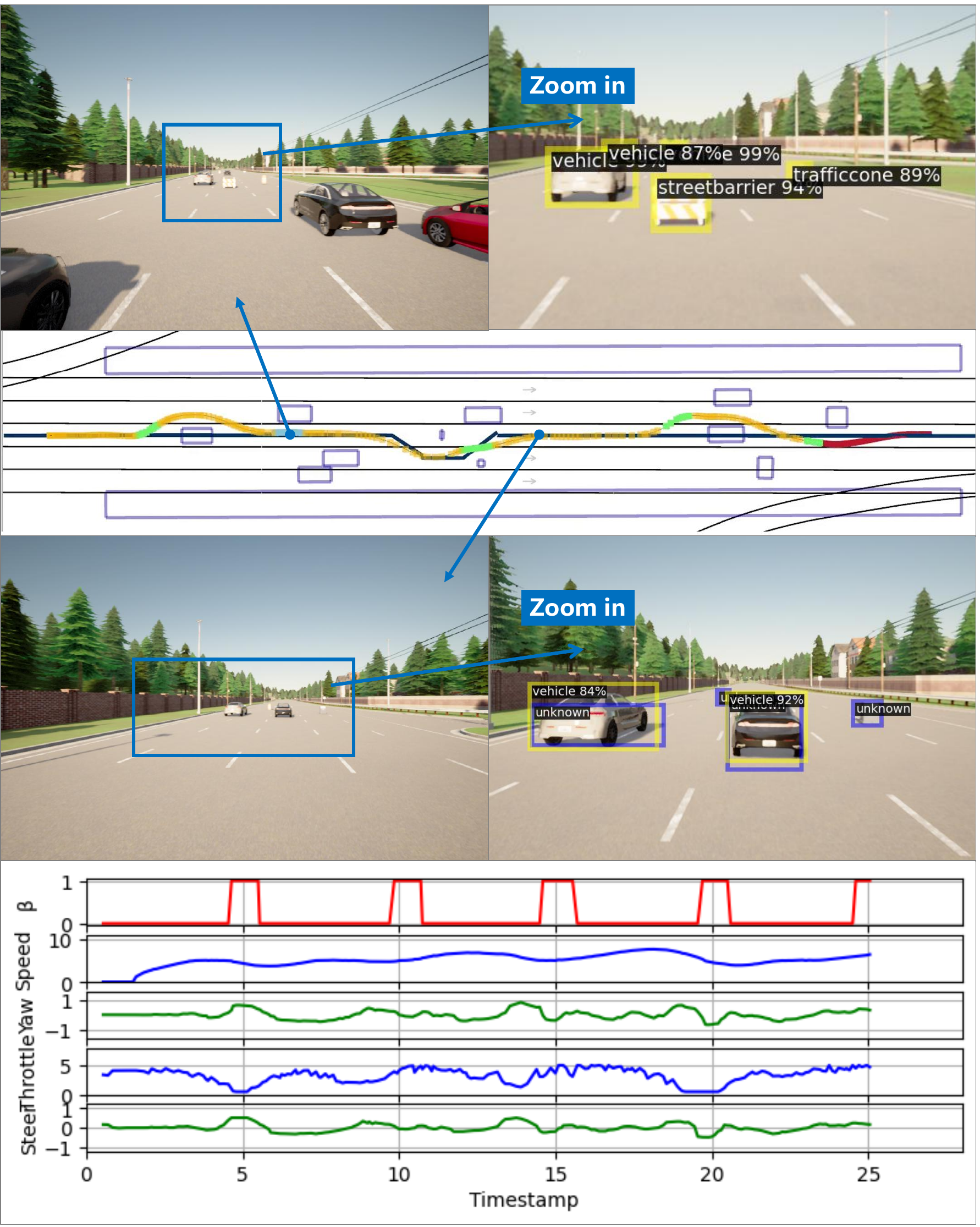}}
    \subfigure[OCP (Ours)]{
    \label{fig:base}
    \includegraphics[width=0.3\textwidth]{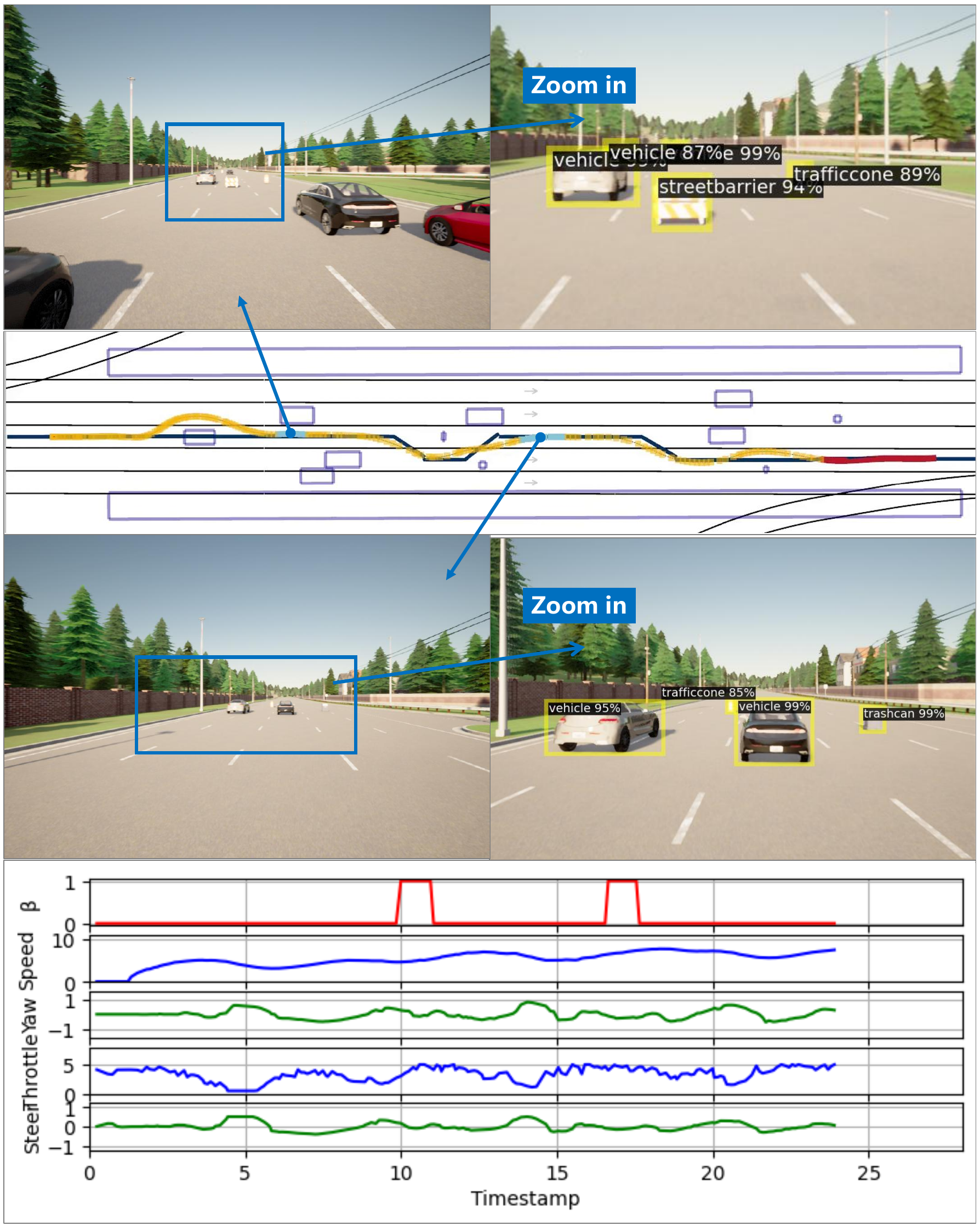}}    
    \vspace{-0.1in}
    \caption{Detections, trajectories, control profiles, and collaboration states of different schemes in Experiment 2.}
    \label{figure:experiment1}
    \vspace{-0.2in}
\end{figure*}

\vspace{-0.1in}
\subsection{Experiment 2: Scenarios with Static Obstacles}

\begin{table}[!t]
\centering
\begin{center}
\caption{Quantitative comparison of Experiment 2.}
\label{figure:params_2}
\vspace{0.05in}
\scalebox{1.1}{
\begin{tabular}{|c|c|c|c|c|c|c|}
\hline
{\textbf{Method}} & {\textbf{FTime}} & {\textbf{TLenh}} & {\textbf{AvgLD}} & {\textbf{SVar}} & {\textbf{MLat}} \\
                            (Unit)
                            & (s)                            & (m)                             & (m)                              & (\text{m/s})                   & (m) \\
                            \hline
LOS  & 25.94 & 124.54 & 0.53 & 1.09 & 1.52 \\
PCS  & 24.09 & 122.4 & 0.59 & 1.13 & 1.89 \\
PCS-2  & 23.53 & 123.84 & 0.57 & 1.39 & 1.69 \\
OCP & 22.35 & 121.85 & 0.6 & 1.3 & 1.76 \\
\hline
\end{tabular}
}
\end{center}
\vspace{-0.3in}
\end{table}

Next, we evaluate the LOS, PCS, and OCP schemes with a target speed of 6 m/s in the Carla Town06 map. 
As shown in Fig.~\ref{figure:setting}, 
the distance between the starting and goal positions is about $110$ meters. 
A total of $7$ known objects (i.e., vehicles) and $4$ unknown objects (i.e., two traffic cones, one streetbarrier, one trashcan) are located between the ego vehicle and its goal. 
These $4$ unknown objects are grouped into two clusters and placed at two regions, forming two challenges within the track.
The trajectories, control profiles, and collaboration states (i.e., $\{\beta_t\}$) of different schemes are shown in Fig.~\ref{figure:experiment1}a--\ref{figure:experiment1}c. 
Yellow trajectories represent local navigation and blue trajectories represent cloud-guided navigation.
Green trajectories represent engagement with large model service but no trajectory change. Red trajectories represent future MPC paths.

Without cloud collaboration, the LOS scheme steers away from all the unknown objects, resulting in the longest finish time and trajectory length. 
The PCS scheme, constrained by fixed time intervals (i.e., 5\,s), passes the first challenge but takes a detour in the second challenge, leading to a trajectory and finish time better than LOS but worse than OCP. The proposed OCP scheme, which dynamically interacts with the cloud based on ODCT and CFS, recognizes and reacts to all unknown obstacles efficiently, achieving the shortest finish time and trajectory length.
More importantly, as seen from Fig.~\ref{figure:experiment1}b and Fig.~\ref{figure:experiment1}c, OCP reduces the number of large model services by $50\%$ compared to PCS (i.e., twice versus $4$ times).
This is achieved by avoiding those improper services at uncritical points (i.e., green parts on the PCS trajectory). 
Enhancing perception at these points would not lead to navigation improvements.
This demonstrates the \textbf{high resource efficiency} brought by the CTO block. 
Interestingly, at the second challenge, the ego-vehicle steers to the wrong direction before receiving the cloud guidance, but soon corrects its path to the right once cloud guidance provides new detections and waypoints. 
This demonstrates the \textbf{self-correction capability} brought by the LVM-MPC block.

\begin{figure}[!t]
    \centering
    \includegraphics[width=0.65\columnwidth]{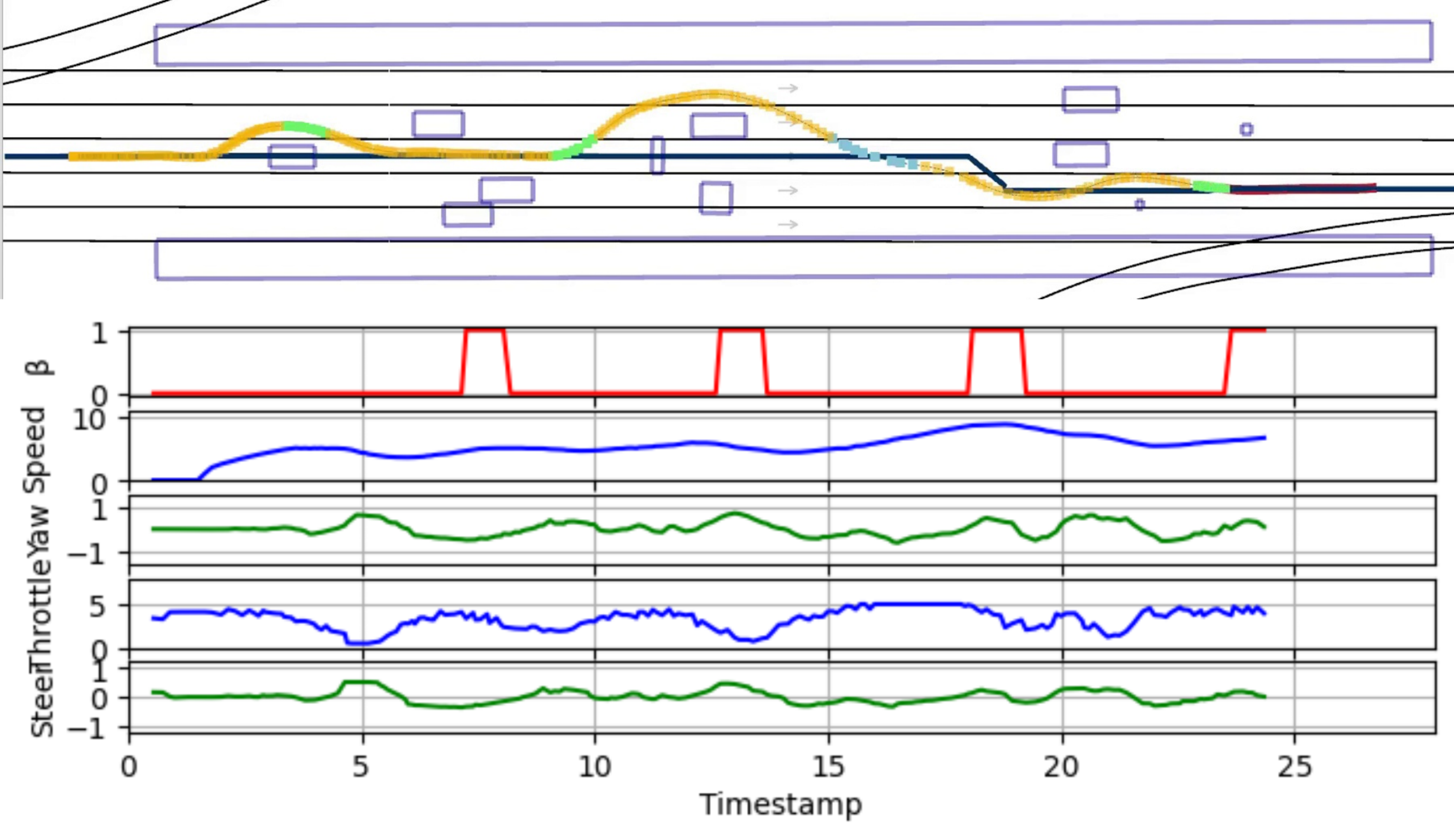}
    \vspace{-0.15in}
    \caption{Trajectory and control profiles of PCS-2.}
    \label{figure:pcs2}
    \vspace{-0.2in}
\end{figure}

Quantitative comparison of the three schemes is shown in Table \ref{figure:params_2}. 
We repeat the experiment 10 times and evaluate the average metrics.
This evaluation consists of finish time (FTime) in s, trajectory length (TLenh) in m, average lateral deviation (AvgLD) in m, speed variability (SVar) in m/s and maximum lateral deviation (MLat) in m.
To ensure a fair comparison, we also simulate a variant PCS scheme, termed PCS-2, which adopts fixed time intervals of 5.5\,s.
Its trajectories and control profiles are shown in Fig. \ref{figure:pcs2}.
It can be seen that the proposed OCP still achieves the the smallest FTime and TLenh, due to the collaboration gain by LVM-MPC and the collaboration timing by CTO. 

\vspace{-0.1in}
\subsection{Experiment 3: More Quantitative Evaluations}

\begin{table}[!t]
\centering
\begin{center}
\caption{Quantitative comparison of Experiment 3.}
\label{table:params}
\vspace{0.05in}
\scalebox{1.0}{
\begin{tabular}{|c|c|c|c|c|c|c|}
\hline
\multirow{2}{*}{$U$} &
{\textbf{Method}} & {\textbf{FTime}} & {\textbf{TLenh}} & {\textbf{AvgLD}} & {\textbf{SVar}} & {\textbf{MLat}} \\
                          &  (Unit)
                            & (s)                            & (m)                             & (m)                              & (\text{m/s})                   & (m) \\
\hline
0 & LOS  & 42.47 & 114.27 & 0.52 & 2.34 & 1.59 \\
0 & PCS  & 42.31 & 114.63 & 0.53 & 2.64 & 1.41 \\
0 & OCP & 43.00 & 113.56 & 0.51 & 2.41 & 1.62 \\
\hline
1 & LOS  & 48.05 & 119.95 & 0.49 & 2.57 & 1.54 \\
1 & PCS  & 44.02 & 114.44 & 0.53 & 2.68 & 1.43 \\
1 & OCP & 44.86 & 113.67 & 0.52 & 2.53 & 1.64 \\
\hline
2 & LOS  & 55.36 & 119.32 & 0.40 & 2.39 & 1.55 \\
2 & PCS  & 47.60 & 114.73 & 0.49 & 2.23 & 1.41 \\
2 & OCP & 45.44 & 112.79 & 0.54 & 2.48 & 1.51 \\
\hline
\end{tabular}
}
\end{center}
\vspace{-0.25in}
\end{table}

\begin{figure}[t]
    \centering
    \noindent
    \includegraphics[width=0.48\columnwidth]{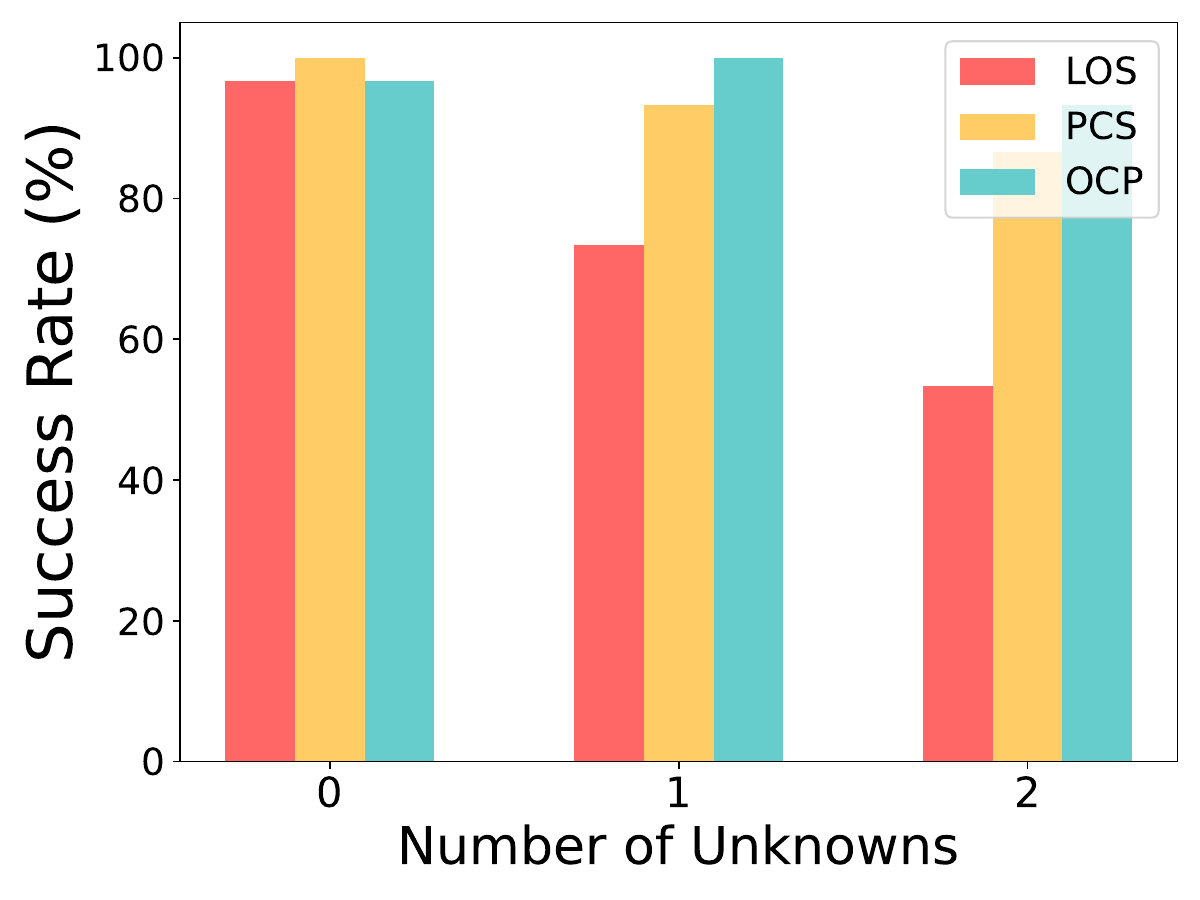}%
    \includegraphics[width=0.48\columnwidth]{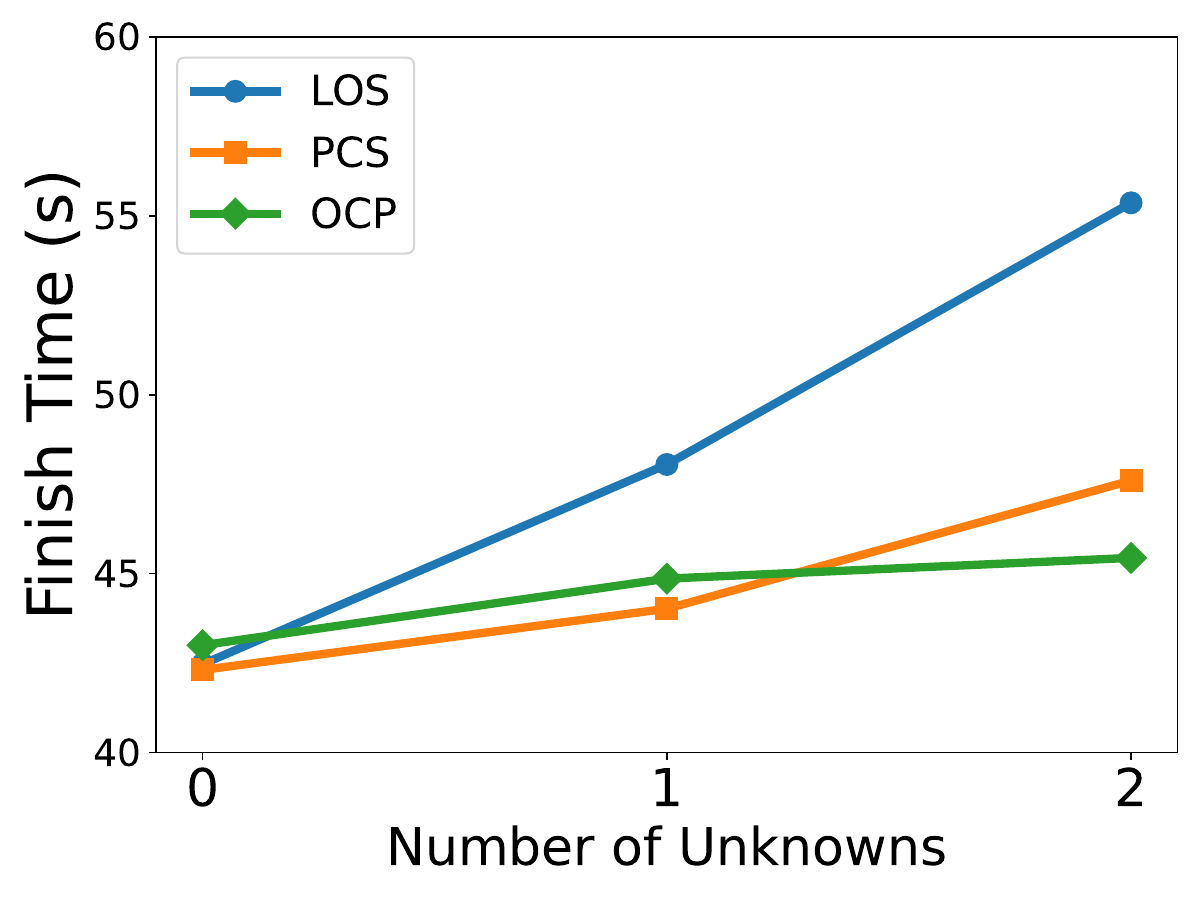}%
    \vspace{-0.15in}
    \caption{Comparison of success rate \& finish time in Exp. 3.}
    \label{figure:experiment1_success}
    \vspace{-0.25in}
\end{figure}

We further evaluate the LOS, PCS, and OCP schemes at a lower speed (target speed of 3 m/s) with smaller horizon ($H = 10$) and time step ($\Delta t=0.1s$) in more scenarios. 
By randomly removing some of the unknown objects in Fig. \ref{figure:setting}, we can configure difference scenarios with $U\in[0, 1, 2]$.
Quantitative results in each scenario are obtained by averaging $30$ random simulation runs, with independent realizations in each run. 
Note that the LOS and PCS schemes may fail due to collision or timeout. 
Here, we only compute their successful cases for evaluations of trajectory qualities, and the failure cases will be used to compute the success rates. 

The evaluation results under different number of unknown objects are shown in Table~\ref{table:params}.
First, as the number of unknown obstacles increases from $0$ to $2$, the trajectory length and finish time significantly increase for all the simulated schemes, due to the necessity of more collision avoidance operations.
Second, the performances of different schemes are similar when the number of unknown objects is zero. However, in presence of a single unknown object, the trajectory length and finish time of LOS increase quickly, as LOS fails in recognizing abnormal objects, leading to either conservative or detouring strategies. 
The PCS scheme overcomes the above challenge aided by the cloud, but the cloud assistance may be invoked at an inappropriate time due to fixed query intervals. 
This leads to performance degradation in presence of multiple unknown objects ($U\geq2$). 
In contrast, the proposed OCP successfully detects all unknown objects using the cloud LVM, while guaranteeing proper query time using CFS, yielding the shortest trajectory length and finish time among all the simulated schemes.

Success rates are computed based on the following two failure conditions: 1) crash failure: ego-vehicle collides with any obstacle or boundary of the lane; 2) no-progress failure: ego-vehicle gets stuck or becomes off the route, failing to find a feasible path.
It can be seen from Fig.~\ref{figure:experiment1_success} the proposed OCP scheme maintains a stable success rate close to $100\,\%$ in all scenarios.
The PCS scheme works well under $0$--$1$ unknown objects, but causes safety issues under $2$ unknown objects.
The LOS scheme leads to the smallest success rates. 
Compared to the second-bast scheme, the proposed OCP reduces the success rate by over $6\%$.

\vspace{-0.35in}
\subsection{Experiment 4: Scenarios with Dynamic Obstacles}

In this experiment, we evaluate our algorithm's performance in a dynamic obstacle scenario at a cross-road in the the Carla Town03 map. The environment includes 2 unknown and 3 known objects. Among the unknowns, one is a static doghouse and the other is a dynamic cybertruck which drives at a constant speed of $4\,$m/s. 
The LOS always fails at the turning point due to the uncertain detection of doghouse and cybertruck, which limits the solution space of MPC. 
Therefore, we only compare the PCS and OCP.

\begin{figure}[!t]
    \centering
    \noindent
    \subfigure[PCS]{
    \label{fig:exp4_pcs}
    \includegraphics[width=0.21\textwidth]{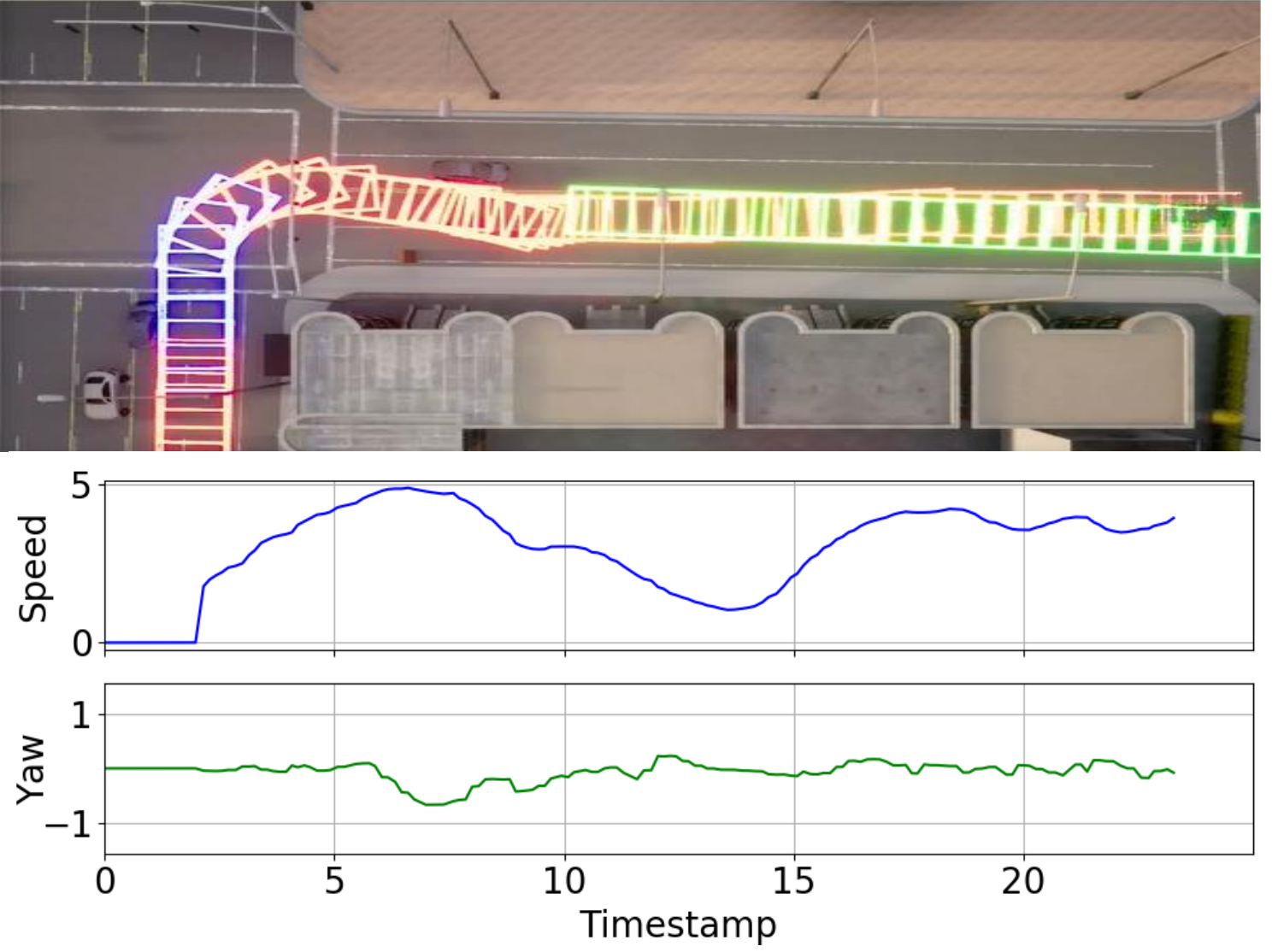}}
    \subfigure[OCP]{
    \label{fig:exp4_ocp}
    \includegraphics[width=0.21\textwidth]{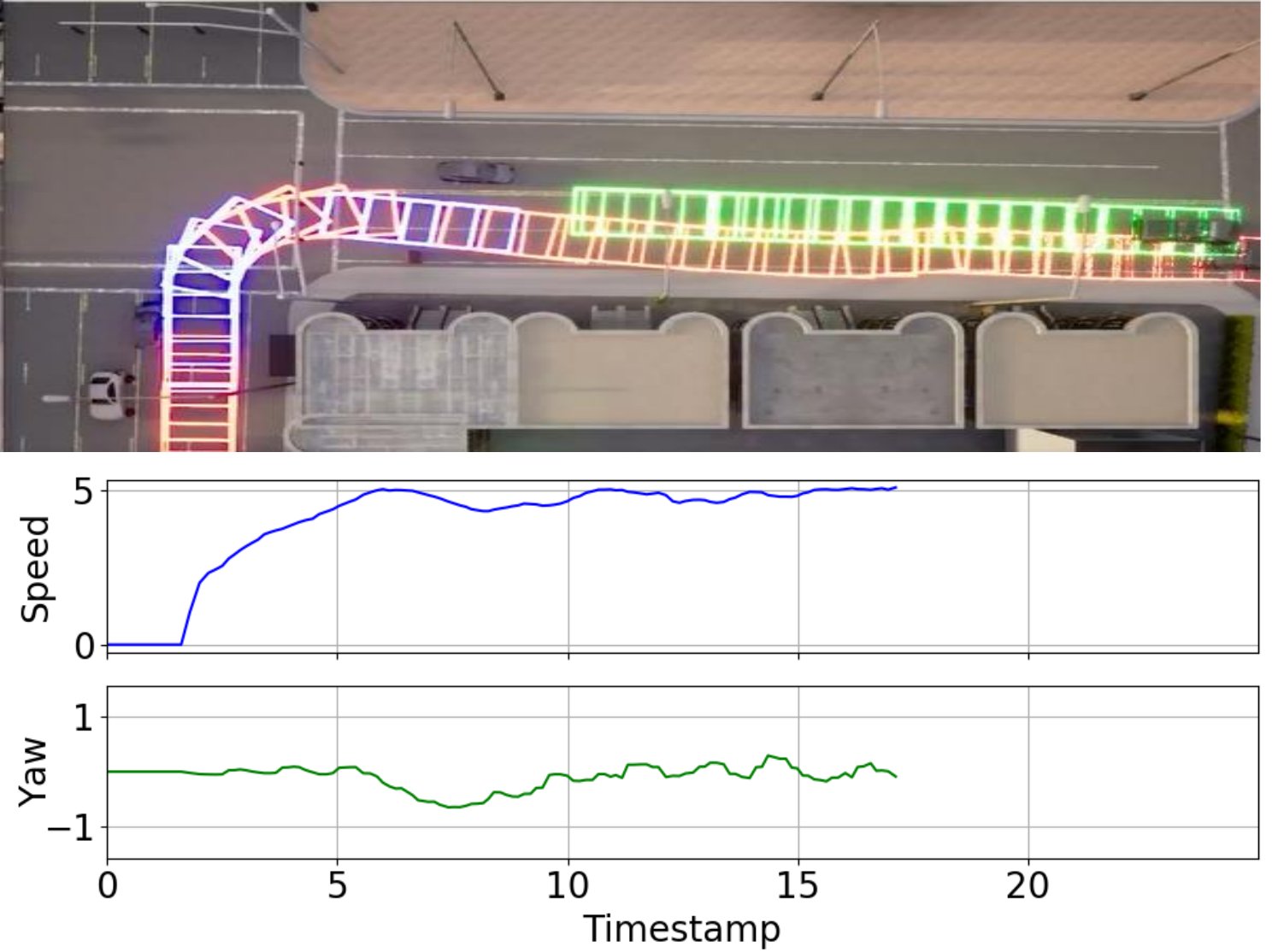}}
    \vspace{-0.1in}
    \caption{Trajectory and control profiles of Experiment 4.}
    \label{figure:experiment4}
    \vspace{-0.2in}
\end{figure}

For PCS, the ego-vehicle turns right while avoiding collision with the doghouse, and follows the lead vehicle cybertruck until reaching the goal. 
In contrast, for OCP, the ego-vehicle triggers another cloud query and service during car following behind the cybertruck. 
This is because CFS rejects the vehicle query at improper timing.
This reserves the chance for subsequent collaboration and make it possible to overtake the cybertruck by following the cloud generated overtaking waypoints rather than the car-following waypoints, as shown in Fig.~\ref{figure:experiment4} (red boxes represent local controller; blue boxes represent LVM-MPC; green boxes represent dynamic obstacle). 
Consequently, compared to PCS, the proposed OCP finishes the trip in a significantly shorter time (over 26.2\% time reduction). This demonstrates the superiority of our OCP strategy in dynamic obstacle scenarios. Details result are shown in Table~\ref{table:exp4}.

\begin{table}[t]
\centering
\begin{center}
\caption{Quantitative comparison of Experiment 4.}
\vspace{0.05in}
\label{table:exp4}
\scalebox{1.1}{
\begin{tabular}{|c|c|c|c|c|c|c|}
\hline
{\textbf{Method}} & {\textbf{FTime}} & {\textbf{TLenh}} & {\textbf{AvgLD}} & {\textbf{SVar}} & {\textbf{MLat}} \\
                            (Unit)
                            & (s)                            & (m)                             & (m)                              & (\text{m/s})                   & (m) \\
                            \hline
PCS   & 23.4                     & 71.47                          & 0.47                              & 2.99                      & 1.68                                 \\ 
OCP    & 17.28                     & 72.90                          & 0.62                              & 2.29          
& 1.44                                 \\ \hline
\end{tabular}
}
\end{center}
\vspace{-0.25in}
\end{table}

\vspace{-0.05in}
\section{Conclusion}
This paper presented OCP, a framework that enhances autonomous navigation in open scenarios by integrating small local models with large cloud models. 
The LVM-MPC collaboration was proposed, which ensures interpretable feasible vehicle actions while enjoying intelligent decisions guided by the cloud.
Furthermore, joint query-service optimization was achieved via CTO, which addresses the problems of when to seek and offer collaboration.
Experiments in the Carla validated the effectiveness of our OCP, in terms of identifying unknown objects and conducting safe efficient autonomous navigation in various scenarios. 
\vspace{-0.05in}
\bibliographystyle{IEEEtran}
\bibliography{main}

\begin{thebibliography}{10}
\providecommand{\url}[1]{#1}
\csname url@samestyle\endcsname
\providecommand{\newblock}{\relax}
\providecommand{\bibinfo}[2]{#2}
\providecommand{\BIBentrySTDinterwordspacing}{\spaceskip=0pt\relax}
\providecommand{\BIBentryALTinterwordstretchfactor}{4}
\providecommand{\BIBentryALTinterwordspacing}{\spaceskip=\fontdimen2\font plus
\BIBentryALTinterwordstretchfactor\fontdimen3\font minus \fontdimen4\font\relax}
\providecommand{\BIBforeignlanguage}[2]{{%
\expandafter\ifx\csname l@#1\endcsname\relax
\typeout{** WARNING: IEEEtran.bst: No hyphenation pattern has been}%
\typeout{** loaded for the language `#1'. Using the pattern for}%
\typeout{** the default language instead.}%
\else
\language=\csname l@#1\endcsname
\fi
#2}}
\providecommand{\BIBdecl}{\relax}
\BIBdecl

\bibitem{dense-rl}
S.~Feng, H.~Sun, X.~Yan, H.~Zhu, Z.~Zou, S.~Shen, and H.~X. Liu, ``Dense reinforcement learning for safety validation of autonomous vehicles,'' \emph{Nature}, vol. 615, no. 7953, pp. 620--627, 2023.

\bibitem{han2024neupan}
R.~Han, S.~Wang, S.~Wang, Z.~Zhang, J.~Chen, S.~Lin, C.~Li, C.~Xu, Y.~C. Eldar, Q.~Hao \emph{et~al.}, ``Neupan: Direct point robot navigation with end-to-end model-based learning,'' \emph{IEEE Transactions on Robotics}, 2025.

\bibitem{han2023rda}
R.~Han, S.~Wang, S.~Wang, Z.~Zhang, Q.~Zhang, Y.~C. Eldar, Q.~Hao, and J.~Pan, ``Rda: An accelerated collision free motion planner for autonomous navigation in cluttered environments,'' \emph{IEEE Robotics and Automation Letters}, vol.~8, no.~3, pp. 1715--1722, 2023.

\bibitem{han2023efficient}
Z.~Han and et~al., ``An efficient spatial-temporal trajectory planner for autonomous vehicles in unstructured environments,'' \emph{IEEE Transactions on Intelligent Transportation Systems}, vol.~25, no.~2, pp. 1797--1814, Oct. 2024.

\bibitem{li2023safe}
D.~Li, B.~Liu, Z.~Huang, Q.~Hao, D.~Zhao, and B.~Tian, ``Safe motion planning for autonomous vehicles by quantifying uncertainties of deep learning-enabled environment perception,'' \emph{IEEE Transactions on Intelligent Vehicles}, vol.~9, no.~1, pp. 2318--2332, Jan. 2024.

\bibitem{kou2025enhancing}
W.-B. Kou, Q.~Lin, M.~Tang, S.~Wang, R.~Ye, G.~Zhu, and Y.-C. Wu, ``Enhancing large vision model in street scene semantic understanding through leveraging posterior optimization trajectory,'' \emph{arXiv preprint arXiv:2501.01710}, 2025.

\bibitem{hu2024agentscodriver}
S.~Hu, Z.~Fang, Z.~Fang, X.~Chen, and Y.~Fang, ``Agentscodriver: Large language model empowered collaborative driving with lifelong learning,'' \emph{arXiv preprint arXiv:2404.06345}, 2024.

\bibitem{xu2023drivegpt4}
Z.~Xu, Y.~Zhang, E.~Xie, Z.~Zhao, Y.~Guo, K.~K. Wong, Z.~Li, and H.~Zhao, ``Drivegpt4: Interpretable end-to-end autonomous driving via large language model,'' \emph{arXiv preprint arXiv:2310.01412}, 2023.

\bibitem{li2024edge}
G.~Li, R.~Han, S.~Wang, F.~Gao, Y.~C. Eldar, and C.~Xu, ``Edge accelerated robot navigation with collaborative motion planning,'' \emph{IEEE/ASME Transactions on Mechatronics}, 2024.

\bibitem{zhang2024multi}
S.~Zhang, H.~Li, S.~Zhang, S.~Wang, D.~W.~K. Ng, and C.~Xu, ``Multi-uncertainty aware autonomous cooperative planning,'' in \emph{2024 IEEE/RSJ International Conference on Intelligent Robots and Systems (IROS)}.\hskip 1em plus 0.5em minus 0.4em\relax IEEE, 2024, pp. 1018--1025.

\bibitem{wang2023bevgpt}
P.~Wang, M.~Zhu, H.~Lu, H.~Zhong, X.~Chen, S.~Shen, X.~Wang, and Y.~Wang, ``Bevgpt: Generative pre-trained large model for autonomous driving prediction, decision-making, and planning,'' \emph{arXiv preprint arXiv:2310.10357}, 2023.

\bibitem{RILaaS}
A.~K. Tanwani, R.~Anand, J.~E. Gonzalez, and K.~Goldberg, ``Rilaas: Robot inference and learning as a service,'' \emph{IEEE Robotics and Automation Letters}, vol.~5, no.~3, pp. 4423--4430, Jul. 2020.

\bibitem{sha2023languagempc}
H.~Sha, Y.~Mu, Y.~Jiang, L.~Chen, C.~Xu, P.~Luo, S.~E. Li, M.~Tomizuka, W.~Zhan, and M.~Ding, ``Languagempc: Large language models as decision makers for autonomous driving,'' \emph{arXiv preprint arXiv:2310.03026}, 2023.

\bibitem{morrell2022nebula}
B.~Morrell, R.~Thakker, {\`A}.~Santamaria~Navarro, A.~Bouman, X.~Lei, J.~Edlund, T.~Pailevanian, T.~S. Vaquero, Y.~L. Chang, T.~Touma \emph{et~al.}, ``{NeBula: TEAM CoSTAR’s robotic autonomy solution that won phase II of DARPA subterranean challenge},'' \emph{Field Robotics}, vol.~2, pp. 1432--1506, 2022.

\bibitem{FogROS2}
J.~Ichnowski and et~al., ``{FogROS2}: An adaptive platform for cloud and fog robotics using {ROS} 2,'' in \emph{IEEE International Conference on Robotics and Automation (ICRA)}, 2023, pp. 5493--5500.

\bibitem{dosovitskiy2017carla}
A.~Dosovitskiy, G.~Ros, F.~Codevilla, A.~Lopez, and V.~Koltun, ``Carla: An open urban driving simulator,'' in \emph{Conference on robot learning}.\hskip 1em plus 0.5em minus 0.4em\relax PMLR, 2017, pp. 1--16.

\bibitem{Kirillov_2023_ICCV}
A.~Kirillov, E.~Mintun, N.~Ravi, H.~Mao, C.~Rolland, L.~Gustafson, T.~Xiao, S.~Whitehead, A.~C. Berg, W.-Y. Lo, P.~Dollar, and R.~Girshick, ``Segment anything,'' in \emph{Proceedings of the IEEE/CVF International Conference on Computer Vision (ICCV)}, October 2023, pp. 4015--4026.

\bibitem{liang2023unknown}
W.~Liang, F.~Xue, Y.~Liu, G.~Zhong, and A.~Ming, ``Unknown sniffer for object detection: Don't turn a blind eye to unknown objects,'' in \emph{Proceedings of the IEEE/CVF Conference on Computer Vision and Pattern Recognition}, 2023, pp. 3230--3239.

\bibitem{ravi2024sam2}
N.~Ravi, V.~Gabeur, Y.-T. Hu, R.~Hu, C.~Ryali, T.~Ma, H.~Khedr, R.~R{\"a}dle, C.~Rolland, L.~Gustafson \emph{et~al.}, ``{Sam 2: Segment Anything in Images and Videos},'' \emph{arXiv preprint arXiv:2408.00714}, 2024.

\bibitem{ding2021epsilon}
W.~Ding, L.~Zhang, J.~Chen, and S.~Shen, ``{EPSILON}: An efficient planning system for automated vehicles in highly interactive environments,'' \emph{IEEE Transactions on Robotics}, vol.~38, no.~2, pp. 1118--1138, 2021.

\bibitem{lauri2016planning}
M.~Lauri and R.~Ritala, ``Planning for robotic exploration based on forward simulation,'' \emph{Robotics and Autonomous Systems}, vol.~83, pp. 15--31, 2016.

\end{thebibliography}

\end{document}